\documentclass[letterpaper, 10 pt, conference]{ieeeconf}

\IEEEoverridecommandlockouts                              

\usepackage{myStyle}

\newacronym[description={Autonomous Surface Vessel: A vessel that can autonomously operate on the surface of water}]{asv}{ASV}{Autonomous Surface Vessel}

\newacronym[description={Target Ship: A dynamic obstacle ship that should be avoided}]{ts}{TS}{Target Ship}

\newacronym[description={Own Ship: The vessel that houses the control system or the perspective from which a situation is being observed}]{os}{OS}{Own Ship}

\newacronym[description={Convention on the International Regulations for Preventing Collisions at Sea: A set of international regulations for preventing collisions between vessels at sea}]{colregs}{COLREGs}{Convention on the International Regulations for Preventing Collisions at Sea}

\newacronym[description={Value at Risk: A statistical measure used to assess the risk which originated from the world of finance}]{var}{VaR}{Value at Risk}

\newacronym[description={Conditional Value at Risk: A risk assessment measure that estimates the expected loss of a process exceeding its VaR}]{cvar}{CVaR}{Conditional Value at Risk}

\newacronym[description={Closest Point of Approach: The nearest point two vessels will come to each other}]{cpa}{CPA}{Closest Point of Approach}

\newacronym[description={Time to Closest Point of Approach: The time remaining until the closest point of approach is reached}]{tcpa}{TCPA}{Time to Closest Point of Approach}

\newacronym[description={Distance to Closest Point of Approach: The distance between the current position of the OS and the CPA}]{dcpa}{DCPA}{Distance to Closest Point of Approach}

\newacronym[description={Collision Risk Index: A numerical indicator of the risk of collision between vessels which is based on the \gls{cpa} metric}]{cri}{CRI}{Collision Risk Index}

\newacronym[description={Automatic Identification System: An automatic tracking system used for identifying and locating vessels by electronically exchanging data with other nearby ships and AIS base stations}]{ais}{AIS}{Automatic Identification System}

\newacronym[description={Automatic Radar Plotting Aid: A system that provides assistance in tracking and plotting the position of other vessels via radar}]{arpa}{ARPA}{Automatic Radar Plotting Aid}

\newacronym[description={Risk Indicator: A measure or metric used to assess the level of risk in a given situation}]{ri}{RI}{Risk Indicator}

\newacronym[description={Time-varying Collision Risk: A metric that can differentiate between situations with identical DCPA and TCPA}]{tcr}{TCR}{Time-varying Collision Risk}

\newacronym[description={Rule-aware Time-varying Collision Risk: An enhanced version of TCR that incorporates navigational rules}]{r-tcr}{R-TCR}{Rule-aware Time-varying Collision Risk}

\newacronym[description={Improved R-TCR: An enhancement to R-TCR that accounts for uncertainties in the \gls{ts}'s trajectory}]{ir-tcr}{Improved R-TCR}{Improved Rule-aware Time-varying Collision Risk}

\newacronym[description={Extended Kalman Filter: An algorithm that extends the Kalman filter to nonlinear systems}]{ekf}{EKF}{Extended Kalman Filter}

\newacronym[description={Instantaneous Collision Probability: The probability of collision at any given instant}]{icp}{ICP}{Instantaneous Collision Probability}

\newacronym[description={Maximum Instantaneous Collision Probability: The highest probability of collision at any moment within a given time horizon}]{micp}{MICP}{Maximum Instantaneous Collision Probability}

\newacronym[description={Deep Neural Network: A complex neural network with multiple layers that can model intricate patterns in data}]{dnn}{DNN}{Deep Neural Network}

\newacronym[description={Artificial Potential Field: A method used in robotics to steer a robot around obstacles by simulating a potential field where obstacles repel and goals attract}]{apf}{APF}{Artificial Potential Field}

\newacronym[description={Velocity Obstacle: A concept used in robotics to describe the set of velocities of a robot that would result in a collision with an obstacle}]{vo}{VO}{Velocity Obstacle}

\newacronym[description={Linear Velocity Obstacles: A simplified version of velocity obstacles that assumes linear paths for both the robot and the obstacle}]{lvo}{LVO}{Linear Velocity Obstacles}

\newacronym[description={Non-Linear Velocity Obstacles: An extension of velocity obstacles that considers non-linear paths for motion planning}]{nlvo}{NLVO}{Non-Linear Velocity Obstacles}

\newacronym[description={Probabilistic Velocity Obstacles: A variation of velocity obstacles that incorporates uncertainty and probabilistic predictions into motion planning}]{pvo}{PVO}{Probabilistic Velocity Obstacles}

\newacronym[description={Generalised Velocity Obstacles: A generalised approach to velocity obstacles that can accommodate different types of motion models and scenarios}]{gvo}{GVO}{Generalised Velocity Obstacles}

\newacronym[description={Finite State Machine: A computational model used to simulate sequential logic and various states within a system}]{fsm}{FSM}{Finite State Machine}

\newacronym[description={Electronic Chart Display and Information System: A computer-based navigation information system that complies with International Maritime Organisation regulations and can be used as an alternative to paper nautical charts}]{ecdis}{ECDIS}{Electronic Chart Display and Information System}

\newacronym[description={Dynamic-Window: A technique used in robotics for obstacle avoidance and navigation}]{dw}{DW}{Dynamic-Window}

\newacronym[description={Adaptive Risk and Contingency-Aware Planner: A lattice-based risk-aware motion planner}]{a-rcap}{A-RCAP}{Adaptive Risk and Contingency-Aware Planner}

\newacronym[description={Deep Reinforcement Learning: An area of machine learning focusing on how agents ought to take actions in an environment to maximise some notion of cumulative reward}]{drl}{DRL}{Deep Reinforcement Learning}

\newacronym[description={Genetic Algorithm: A search heuristic that mimics the process of natural selection to generate high-quality solutions to optimisation and search problems}]{ga}{GA}{Genetic Algorithm}

\newacronym[description={Evolutionary Algorithm: A subset of evolutionary computation, a generic population-based metaheuristic optimisation algorithm}]{ea}{EA}{Evolutionary Algorithm}

\newacronym[description={Model Predictive Control: A type of control algorithm that uses a model to predict future outcomes and optimise control actions}]{mpc}{MPC}{Model Predictive Control}

\newacronym[description={Deterministic MPC: A variant of MPC that operates under deterministic assumptions about the system and its environment}]{dmpc}{DMPC}{Deterministic MPC}

\newacronym[description={Stochastic MPC: A form of MPC that considers stochastic elements and uncertainties in the system model}]{smpc}{SMPC}{Stochastic MPC}

\newacronym[description={Robust MPC: A type of MPC designed to maintain performance in the presence of bounded disturbances and uncertainties}]{rmpc}{RMPC}{Robust MPC}

\newacronym[description={Scenario-Based MPC: An approach in MPC that deals with uncertainties by considering different future scenarios}]{sb-mpc}{SB-MPC}{Scenario-Based MPC}

\newacronym[description={Model Predictive Path Integral control: A control method that combines MPC with path integral control theory in the form of a sampling based MPC}]{mppi}{MPPI}{Model Predictive Path Integral control}

\newacronym[description={Kullback-Leibler divergence: A measure of how one probability distribution diverges from a second, expected probability distribution}]{kl}{KL}{Kullback-Leibler}

\newacronym[description={Unmanned Aerial Vehicle: An aircraft operated without a human pilot on board}]{uav}{UAV}{Unmanned Aerial Vehicle}

\newacronym[description={Risk-Aware Model Predictive Path Integral: An extension to MPPI where uncertainties in the dynamics of the controlled vehicle are accounted for}]{ra-mppi}{RA-MPPI}{Risk-Aware Model Predictive Path Integral}

\newacronym[description={Probabilistic Scenario-Based MPC: A control approach that uses probabilistic scenarios to predict and optimise the behavior of a system}]{psb-mpc}{PSB-MPC}{Probabilistic Scenario-Based MPC}

\newacronym[description={Smooth Model Predictive Path Integral: An extension to MPPI where no extra smoothening step is required after sampling}]{smppi}{SMPPI}{Smooth Model Predictive Path Integral}

\newacronym[description={Ensemble Model Predictive Path Integral: An extension to MPPI where multiple possible scenarios are optimised by multiple MPPIs}]{emppi}{EMPPI}{Ensemble Model Predictive Path Integral}

\newacronym[description={Covariance-Controlled Model Predictive Path Integral: An extension to MPPI where the sampling distribution is controlled using Covariance Steering}]{cc-mppi}{CC-MPPI}{Covariance-Controlled Model Predictive Path Integral}

\newacronym[description={Covariance Steering: A technique in control theory focusing on steering a system's state while considering its variance or uncertainty}]{cs}{CS}{Covariance Steering}

\newacronym[description={Discrete-Time Control Barrier Functions: Functions used in control systems to ensure safety constraints are met over time}]{dcbf}{DCBF}{Discrete-Time Control Barrier Functions}

\newacronym[description={Adaptive Importance Sampling: A statistical technique for efficiently estimating properties of a particular distribution, by adapting the sampling technique based on previous samples}]{ais2}{AIS}{Adaptive Importance Sampling}

\newacronym[description={Center Of Gravity: The point in a body or system around which its mass or weight is evenly distributed or balanced}]{cog}{COG}{Center Of Gravity}

\newacronym[description={Dynamic Risk-Aware Model Predictive Path Integral control: Our proposed risk-aware motion planner which is an extension to MPPI}]{dra-mppi}{DRA-MPPI}{Dynamic Risk-Aware Model Predictive Path Integral control}

\newacronym[description={Spatial Collision Risk: A metric or system used to assess the risk of collision for ship domains}]{scr}{SCR}{Spatial Collision Risk}

\newacronym[description={Fuzzy Collision Danger Domain: A ship domain that adapts to the conditions and changes shape based on fuzzy logic}]{fcdd}{FCDD}{Fuzzy Collision Danger Domain}

\newacronym[description={Quaternion Ship Domain: A analytical domain created by specifying different shapes for each quaternion around the ship}]{qsd}{QSD}{Quaternion Ship Domain}

\newacronym[description={Last Line of Defence: A type of action lines approach}]{llod}{LLoD}{Last Line of Defence}

\newacronym[description={Neural Network: A computational model inspired by the human brain, used in machine learning for pattern recognition and decision making}]{nn}{NN}{Neural Network}

\newacronym[description={Variance of Compass Degree: A measure of the variability or inconsistency in the readings of a compass}]{vcd}{VCD}{Variance of Compass Degree}

\newacronym[description={Vehicle Traffic Service: A service designed to monitor and manage vehicle traffic, especially in crowded or complex areas}]{vts}{VTS}{Vehicle Traffic Service}

\newacronym[description={Evidential Reasoning: A decision-making process based on evaluating evidence and determining its reliability and relevance}]{er}{ER}{Evidential Reasoning}

\newacronym[description={Rapidly exploring Random Tree: An algorithm for efficiently searching high-dimensional spaces by randomly building a space-filling tree}]{rrt}{RRT}{Rapidly exploring Random Tree}

\newacronym[description={Sum-Of-Squares: A mathematical approach that involves summing the squares of variables or terms, often used in optimisation and analysis}]{sos}{SOS}{Sum-Of-Squares}

\newacronym[description={Fuzzy Case-Based Reasoning: A method combining fuzzy logic with case-based reasoning to deal with uncertain or imprecise information}]{fcbr}{FCBR}{Fuzzy Case-Based Reasoning}

\newacronym[description={Business Intelligence and Strategy: A company that does research and specialises in market intelligence on Deep Tech}]{bis}{BIS}{Business Intelligence and Strategy}

\newacronym[description={European Maritime Safety Agency: An agency of the European Union charged with reducing the risk of maritime accidents, marine pollution, and loss of human life at sea}]{emsa}{EMSA}{European Maritime Safety Agency}

\newacronym[description={Massachusetts Institute of Technology: A prestigious private research university in Cambridge, Massachusetts, known for its advanced research and education in technology and science}]{mit}{MIT}{Massachusetts Institute of Technology}

\newacronym[description={Amsterdam Institute for Advanced Metropolitan Solutions: An institute focused on urban planning and technology solutions for metropolitan areas, based in Amsterdam}]{ams}{AMS}{Amsterdam Institute for Advanced Metropolitan Solutions}

\newacronym[description={International Maritime Organisation: A specialised agency of the United Nations responsible for regulating shipping.}]{imo}{IMO}{International Maritime Organisation}

\newacronym[description={Binnenvaart Politie Regelement: A set of regulations for preventing collisions between vessels in the Dutch inland waters}]{bpr}{BPR}{Binnenvaart Politie Regelement}

\newacronym[description={Interaction-Aware MPPI: a method developed to coordinate navigation between ASVs in urban canals.}]{ia-mppi}{IA-MPPI}{Interaction-Aware MPPI}

\newacronym[description={Collision Probability: the probability that two agents collide}]{cp}{CP}{Collision Probability}

\newacronym[description={DRA-MPPI Compliant: an ablated version of \gls{dra-mppi} that has no robustness against non-compliant agents.}]{dra-mppi-c}{DRA-MPPI-C}{DRA-MPPI Compliant}

\newacronym[description={DRA-MPPI Non Compliant: an ablated version of \gls{dra-mppi} that has no rule awareness.}]{dra-mppi-nc}{DRA-MPPI-NC}{DRA-MPPI Non Compliant}

\newacronym[description={Monte Carlo: A computational algorithm that uses random sampling to approximate mathematical functions or simulate processes}]{mc}{MC}{Monte Carlo}

\begin{document}

\title{\LARGE \bf
Dynamic Risk-Aware MPPI for Mobile Robots in Crowds\\via Efficient Monte Carlo Approximations
}

\author{Elia Trevisan$^{1*}$, Khaled A. Mustafa$^{1*}$, Godert Notten$^{1,2}$, Xinwei Wang$^{3}$, and Javier Alonso-Mora$^1$ 
\thanks{This research is supported by the project ``Sustainable Transportation and Logistics over Water: Electrification, Automation and Optimization (TRiLOGy)'' of the Netherlands Organization for Scientific Research (NWO), domain Science (ENW), the Amsterdam Institute for Advanced Metropolitan Solutions (AMS) in the Netherlands, and the Dutch Research Council NWO-NWA, within the “Acting under uncertainty” (ACT) project (Grant No. NWA.1292.19.298).}
\thanks{$^*$These authors have contributed equally.}
\thanks{$^{1}$Cognitive Robotics Department,
        TU Delft, 
        {\tt\small \{e.trevisan, k.a.mustafa, j.alonsomora\}@tudelft.nl}}
\thanks{$^{2}$Damen Naval}
\thanks{$^{3}$Centre for Intelligent Transport, Queen Mary University of London,
        {\tt\small xinwei.wang@qmul.ac.uk}}}
\maketitle
\thispagestyle{empty}
\pagestyle{empty}

\begin{abstract}
Deploying mobile robots safely among humans requires the motion planner to account for the uncertainty in the other agents' predicted trajectories. This remains challenging in traditional approaches, especially with arbitrarily shaped predictions and real-time constraints.
To address these challenges, we propose a Dynamic Risk-Aware Model Predictive Path Integral control (DRA-MPPI), a motion planner that incorporates uncertain future motions modelled with potentially non-Gaussian stochastic predictions. By leveraging MPPI’s gradient-free nature, we propose a method that efficiently approximates the joint Collision Probability (CP) among multiple dynamic obstacles for several hundred sampled trajectories in real-time via a Monte Carlo (MC) approach.
This enables the rejection of samples exceeding a predefined CP threshold or the integration of CP as a weighted objective within the navigation cost function. Consequently, DRA-MPPI mitigates the freezing robot problem while enhancing safety. Real-world and simulated experiments with multiple dynamic obstacles demonstrate DRA-MPPI’s superior performance compared to state-of-the-art approaches, including Scenario-based Model Predictive Control (S-MPC), Frenét planner, and vanilla MPPI.
Videos of the experiments can be found at \href{https://autonomousrobots.nl/paper_websites/dra-mppi}{https://autonomousrobots.nl/paper\_websites/dra-mppi}.
\end{abstract}


\section{Introduction}
Mobile robots have the potential to enhance various aspects of daily life, from optimizing logistics in warehouses to enabling safer and more efficient transportation through autonomous vehicles \cite{amazon, SD, MI}. However, for robots to be successfully integrated into real-world settings like urban areas, they must be capable of safely and efficiently manoeuvring through human-populated spaces. Achieving this requires an ability to interpret and anticipate human movement—a task complicated by the inherent unpredictability of human behaviour. Prediction models, such as \cite{pfeiffer}, provide probabilistic distributions over potential human trajectories. To ensure safe and efficient navigation, these probabilistic predictions must be incorporated into the motion planning process \cite{risk-aware, GP, STL}.

A key challenge in planning under uncertainty is finding a safe trajectory despite the stochastic nature of surrounding obstacles. One approach to handle uncertainty in dynamic environments is \textit{robust optimization} \cite{robust-optimization}, which enforces safety guarantees by considering worst-case scenarios within bounded uncertainty sets. This method assumes that the probability density of uncertainty is nonzero within a defined region of the ego agent’s workspace, ensuring strict safety constraints. However, by accounting for all possible uncertainty realizations, robust optimization often leads to overly cautious behaviour, potentially rendering solutions infeasible in densely populated environments—a phenomenon commonly referred to as the ``frozen robot" problem \cite{Freezing}. To mitigate excessive conservatism, \textit{stochastic optimization} offers an alternative by employing chance constraints \cite{zhu_chance-constrained_2019, CC, de_groot_scenario-based_2021} that probabilistically bound the likelihood of constraint violations within a specified confidence level $\sigma$. This relaxation allows for more flexible decision-making while maintaining a controlled level of risk, enabling robots to navigate complex environments more effectively than purely robust approaches. 
In this work, we propose a risk-aware Model Predictive Path Integral (MPPI) that uses Monte-Carlo sampling to approximate the chance constraints for a mobile robot navigating among dynamic agents.

\section{Related Work} \label{sec:relatedworks}
In stochastic optimization, constraint violations are regulated using chance constraints, which ensure that the probability of satisfying a given nominal constraint remains within a predefined threshold. Evaluating chance constraints exactly is often infeasible in real-world scenarios, necessitating approximations. Many existing methods address this challenge by introducing simplifying assumptions, such as modelling uncertainty with Gaussian distributions \cite{zhu_chance-constrained_2019} or restricting the robot's dynamics to linear systems \cite{Okamoto, Hewing}. While these approximations facilitate tractability, they also limit the flexibility of the approach. Recent research has made significant progress in overcoming these limitations, enabling stochastic optimization to accommodate nonlinear robot dynamics and more general uncertainty distributions \cite{de_groot_scenario-based_2021} by leveraging scenario optimization \cite{SP}. However, the Collision Probability (CP) in these works is computed per obstacle, which degrades performance when
more obstacles influence the plan, i.e., in crowded environments. In addition, empirical evaluations indicate that the \textit{actual risk} associated with the planned trajectory is significantly lower than the upper bound imposed by the chance constraint formulation \cite{zhu_chance-constrained_2019, CC, de_groot_scenario-based_2021}. To address this issue, \cite{mustafa} introduced a method to quantify the risk of planned trajectories by running multiple probabilistic planners in parallel with varying risk thresholds.  This approach provides probabilistic safety guarantees by attaining a closer bound to the specified risk. However, the scalability of this approach is constrained by the number of probabilistic planners that can be executed in parallel.
With the advantage of not requiring gradient information, probabilistic safety measures have been successfully employed in sampling-based motion planning frameworks. In \cite{Luders, Aoude}, chance constraints are applied to the Rapidly Random Trees (RRT) algorithm such that each node in the tree is statistically safe. \cite{nyberg_risk-aware_2021} integrates risk measures to estimate the risk of violating a predefined safety
specification into a sampling-based trajectory \cite{Frenet} to plan minimal risk trajectories.
Model Predictive Path Integral (MPPI), a gradient-free local motion planner, has the advantage of massive parallelizability to compute finite horizon plans online via sampling~\cite{williams_information-theoretic_2018}. 
\cite{yin_risk-aware_2022} integrates the Conditional Value-at-Risk (CVaR) measure into MPPI to generate optimal control actions but only accounts for the uncertainty in the robot's own dynamics.

In this paper, we propose Dynamic Risk-Aware MPPI (DRA-MPPI), a planner approximating joint CPs among several dynamic obstacles by evaluating the probabilistic distributions of dynamic agents' predicted trajectories, which can be non-Gaussian. The approximated probability of collision becomes a chance constraint by assigning a high enough cost to samples that violate the maximum threshold, essentially rejecting them from the final plan. Leveraging Monte Carlo (MC) sampling \cite{caflisch_monte_1998} enables greater flexibility and computational efficiency, as it effectively handles arbitrary probability distributions and scales well with an increasing number of dynamic agents. 

\subsection{Contributions}\label{sec:contribution}
Our main contributions are twofold:
\begin{enumerate}
\item DRA-MPPI: An MPPI-based planner that approximates the CPs of trajectories using MC methods. It handles joint CPs over several dynamic obstacles and manages non-Gaussian predictions with minimal approximations.
\item A computationally efficient strategy to find the MC approximation of the joint CP for several hundred samples in parallel, achieving real-time performance.
\end{enumerate}

We design a comprehensive cost function for efficient navigation and validate the approach through simulations and real-world experiments with pedestrians. Our method, tested with Gaussian and non-Gaussian models, is compared to a state-of-the-art Scenario-based Model Predictive Control (S-MPC), a risk-aware Frenét planner, and a vanilla MPPI.

\section{Preliminaries}\label{sec:preliminaries}
We consider a controlled robot moving in a 2D plane, $\mathcal{W} \subseteq \mathbb{R}^2$, with non-linear discrete-time dynamics given by:
\begin{equation}
    \boldsymbol{x}_{t+1} = \mathcal{F}(\boldsymbol{x}_t, \boldsymbol{u}_t),
\end{equation}
where $\boldsymbol{x}_t = \left[\boldsymbol{q}_t, \psi_t\right] \in \mathbb{R}^{n_x}$ and $\boldsymbol{u}_t \in \mathbb{R}^{n_u}$ denote the state and control input of the robot at stage $t$ respectively. The state of the robot $\boldsymbol{x}_t$ contains its position $\boldsymbol{q}_t = [x_t,y_t] \in \mathbb{R}^2 \subseteq \mathbb{R}^{n_x}$ and orientation $\psi_t$. The robot navigates in an environment occupied by $N_o$ dynamic obstacles whose future trajectories must be predicted. This necessitates a dynamic model of the obstacles' motion, enabling the robot to proactively account for the obstacles' behaviour while generating its own trajectory.

\subsection{Dynamic Obstacle Model}
The motion planner uses a model that probabilistically forecasts the behaviour of surrounding agents. In this setting, we model the positions of $N_o$ dynamic obstacles as random variables. Specifically, the uncertain position of obstacle $o$ at time step $t$ as $\boldsymbol{\delta}^o_t = [x_t^o, y_t^o] \in \Delta$, where $\Delta$ is the probability space capturing the uncertainty in the perception of these obstacles. This uncertainty is characterized by a probability measure $\mathbb{P}$. 
To capture the inherent multi-modality in obstacle behaviour, we model the uncertainty in their motion using a Mixture-of-Gaussians (MoG) distribution. This approach provides a flexible and expressive representation of uncertainty by combining multiple continuous probability distributions.  Specifically, the probability density function of an obstacle’s position at time step $t$ is given by:
\begin{equation}
    p_t^o(x,y) = \sum_{i=1}^{N_m}\phi_i p_{t,i}^o(x,y),
\end{equation}
where $N_m$ is the number of modes of the MoG, $\phi_i$ represents the weight associated with each mode such that $\sum_{i=1}^{N_m} \phi_i= 1$, and $p_{t,i}^o(\cdot)$ is the probability density function of the $i^\text{th}$ mode with mean $\boldsymbol{\mu}_i \in \mathbb{R}^2$ and covariance $\boldsymbol{\Sigma}_i \in \mathbb{R}^{2 \times 2}$:
\begin{equation}
    p_{t,i}^o(\cdot) = \mathcal{N}(\boldsymbol{\mu}_i(t), \boldsymbol{\Sigma}_i(t)).
\end{equation}
As a result, the prediction model outputs a sequence of state distributions over the prediction horizon for each possible mode, enabling the planner to account for the stochastic nature of obstacle motion.

\subsection{Probabilistic Collision Avoidance}
The problem of planning under uncertainty can be formulated as a chance-constraint collision avoidance problem. Given a cost function $J$, the initial state of the robot $\boldsymbol{x}_0 = \boldsymbol{x}_\text{\textnormal{init}}$, and the predicted future states of all obstacles $p_{t,i}^o(\cdot)$ over a horizon $T$, the objective is to compute optimal control inputs that guide the robot from its initial state to progress along a reference path, while the collision probability with any moving obstacle at each stage $t$ is below an acceptable threshold $\sigma$. The resulting optimization problem is given by:
\begin{subequations}
\begin{alignat}{2}
\min_{\boldsymbol{u} \in \mathbb{U}} \quad & \sum_{t=0}^{T-1}J_t(\boldsymbol{x}_t, \boldsymbol{u}_t) + J_T(\boldsymbol{x}_T)\\
\textrm{s.t.} \quad & \boldsymbol{x}_0 = \boldsymbol{x}_{\text{init}},\\
 & \boldsymbol{x}_{t+1} = \mathcal{F}(\boldsymbol{x}_t, \boldsymbol{u}_t), \quad \boldsymbol{x} \in \mathbb{X}, \boldsymbol{u} \in \mathbb{U}, \label{eq1b}\\
  &\mathbb{P} \left[||\boldsymbol{q}_t - \boldsymbol{\delta}_t^o||_2 > r, \forall o\right] \geq 1-\sigma, \forall t, \label{CC}  
\end{alignat}
\end{subequations}
where $J_t(\boldsymbol{x}_t, \boldsymbol{u}_t)$ represents the stage cost of the robot, and $J_T(\boldsymbol{x}_T)$ denotes the terminal cost. States $\boldsymbol{x}_t$ and inputs $\boldsymbol{u}_t$ are bounded by the state and input constraint sets $\mathbb{X}$ and $\mathbb{U}$, and radius $r$ is the sum of the robot and obstacle radii.
\subsection{Model Predictive Path Integral Control (MPPI)}
Model Predictive Path Integral control (MPPI) \cite{williams_information-theoretic_2018, mppi_williams} is a sampling-based optimization method for solving stochastic optimal control problems. We consider a discrete-time dynamical system governed by the state-transition equation:
\begin{equation}
    \boldsymbol{x}_{t+1}=\mathcal{F}(\boldsymbol{x}_t, \boldsymbol{v}_t) \quad \boldsymbol{v}_t \sim \mathcal{N}(\boldsymbol{u}_t, \Sigma),
\end{equation}
where $\mathcal{F}$ represents a nonlinear state-transition function that dictates the evolution of the state  $\boldsymbol{x}$ over discrete time steps $t$. The control input $\boldsymbol{v}_t$ follows a Gaussian distribution centered at the commanded input $\boldsymbol{u}_t$ with covariance
$\Sigma$. To generate potential system trajectories, a set of $K$ perturbed control sequences sequences $V_k$ is sampled and rolled-out through the model $\mathcal{F}$, producing $K$ corresponding state trajectories $X_k, k \in [0, K-
1]$, over a planning horizon of length $T$.  Given the state trajectories $X_k$ and a predefined cost function $\mathcal{C}$ to be minimized, the total state-cost $S_k$
for each sampled sequence $V_k$ is computed by evaluating $S_k = \mathcal{C}(X_k)$. An importance-sampling weighting scheme is then employed to approximate the optimal control sequence. The weight $w_k$ associated with trajectory $X_k$ is computed using an inverse exponential function of the cost $S_k$, adjusted by a tuning parameter $\beta$ known as the inverse temperature:
\begin{equation}
    w_k = \frac{1}{\eta} \exp \left(-\frac{1}{\beta}(S_k -\rho)\right), \quad \sum_{k=1}^{K}w_k =1. \label{IS}
\end{equation}
Here, the minimum sampled cost $\rho = \min_k S_k$ is introduced to enhance numerical stability, and $\eta$ is a normalization factor ensuring weights sum to unity.  The final control input sequence $U^*$ is then computed as a weighted sum of the sampled control sequences:
\begin{equation}\label{eq:mppi_update}
    U^* = \sum_{k=1}^K w_kV_k.
\end{equation}
Once the optimal control sequence $U^*$ is obtained, only the first control input $\boldsymbol{u}_0^*$ is applied to the system. The procedure is then repeated at the next time step, where, at the next iteration, a time-shifted $U^*$ is used as a warm-start.

\subsection{Risk Assessment}
To quantify risk in the context of pedestrian-robot interactions, we define a risk metric that assesses the likelihood of collision. Let $ \mathcal{Z} $ represent the set of random variables capturing the uncertainty in pedestrian motion in the $x$ and $y$ directions. The risk metric is a function $\zeta: \mathcal{Z} \mapsto \mathbb{R}$ that maps these random variables to a real value corresponding to the collision probability. The probability of collision with an obstacle $o$, the \textit{marginal} probability, is then expressed as:
\begin{equation}
    \mathcal{P}_t^{o}(\boldsymbol{q}_t) =
    \iint_{x_t, y_t \in \mathcal{R}^\circ_t}  p_t^{o}(x_t,y_t) dx dy, \forall{t}, o, \label{eq:cp}
\end{equation}
where the integration domain $\mathcal{R}^\circ_t$ is defined as a circular region, with radius $r$, centered at the planned robot pose $\boldsymbol{q}_t$ at stage $t$ along the planning horizon.

When $o>1$, the probability of colliding with at least one obstacle is given by the \textit{joint} collision probability~\cite{de_groot_scenario-based_2023}. Since, at each time step $t$, the \textit{marginal} probabilities can be assumed independent, the \textit{joint} collision probability can be computed using a multiplicative approach:
\begin{equation}\label{eq:joint_cp}
\mathcal{P}_t(\boldsymbol{q}_t) = 1 - \prod_o^{N_o} (1 - \mathcal{P}_t^{o}(\boldsymbol{q}_t))
\end{equation}

\section{Proposed Approach}
To assess the risk associated with the robot's optimal trajectory in MPPI, \cref{eq:cp,eq:joint_cp} must be evaluated for each sample and each time step in the horizon. For example, with $K=400$ MPPI samples and a horizon $T=20$, we must compute the collision probability $8000$ times at each controller iteration. Computing these analytically is not feasible if the controller is to run in real-time. In \Cref{sec:MC}, we propose a Monte Carlo approximation to estimate the \textit{joint} collision probability for many trajectories efficiently and in parallel. The cost function used to evaluate MPPI samples is detailed in \Cref{sec:cost}, and an overview of the algorithm is presented in \Cref{sec:algo}.
\subsection{Monte Carlo Approximation}
\label{sec:MC}
We propose an algorithm to compute all $N_o$ integral approximations for all $K$ samples efficiently and in parallel, allowing good scalability with the number of dynamic obstacles and samples. The process begins by sampling $N_{mc}$ Monte Carlo coordinates from a rectangular collision region $\mathcal{R}^{{\scriptscriptstyle\square}}_t$, which is bounded as follows:
\begin{equation}\label{eq:sampling_region}
        \mathcal{R}^{{\scriptscriptstyle\square}}_t=\{(x,y) \mid x_\text{lower} \leq x \leq x_\text{upper}, y_\text{lower} \leq y \leq y_\text{upper} \}.
\end{equation}
The bounds of this region are determined by the minimum and maximum spatial positions of the robot's states across all rollouts $K$ at time step $t$, extended by the radius $r$:
\[
    x_\text{lower} = \min_k (x_{k,t}) - r, \quad 
    x_\text{upper} = \max_k (x_{k,t}) + r
\]
\[
    y_\text{lower} = \min_k (y_{k,t}) - r, \quad 
    y_\text{upper} = \max_k (y_{k,t}) + r.
\]
Here, $x_{k,t}$ and $y_{k,t}$ denote the $x$ and $y$ robot's position for rollout $k$ at timestep $t$. By taking the
minimum and maximum values as bounds and extending the area with the integral radius $r$, we cover
the whole area where we need to approximate the integrals at time step $t$.\\
Depending on the obstacles' trajectory predictions, we get $N_o \times N_m$ total number of modes for all obstacles. To approximate the integral for all rollouts $K$ we first evaluate the \textit{joint} collision probability for all MC sample $j$:
\begin{equation}\label{eq:jointCP_persample}
    \mathcal{P}^\text{joint}_{j}(x_j,y_j) = 1 - \prod_{o=1}^{N_o} (1-p_t^o(x_j, y_j)), \quad \forall j
\end{equation}
where $x_j, y_j$ represents a single MC sample. We then identify which of these sampled coordinates fall within the region $\mathcal{R}^{\circ}_{k,t}$ of each $k$ and compute the CP over the relevant domain:
\begin{equation}
\hat{\mathcal{P}}(x_{k,t}, y_{k,t}) = \frac{\pi r^2}{N_{\mathcal{R}^{\circ}_{k,t}}} \sum_{j=1}^{N_{mc}}  \mathcal{P}^{\text{joint}}(x_j,y_j)\cdot\mathds{1}_{\mathcal{R}^{\circ}_{k,t}}(x_j, y_j), \label{mc}
\end{equation}
where $\pi r^2$ is the area of the collision region and $N_{\mathcal{R}^{\circ}_{k,t}}$ is the number of Monte Carlo samples in the circular region $\mathcal{R}^{\circ}_{k,t}$ acting as a normalization factor. $\mathds{1}_{\mathcal{R}^{\circ}_{k,t}}$ is an indicator function to consider only MC samples in the circular collision region $\mathcal{R}^{\circ}_{k,t}$.
Given that all the coordinates $x_{k,t}$ and $y_{k,t}$ will be fairly close $\forall k$, there will also be significant overlap in the integration regions $\mathcal{R}^{\circ}_{k,t}$.
Thus, the computation effort is massively reduced by using our approach of drawing and evaluating the MC samples only once $\forall k$.
The Standard Error (SE) of $\hat{\mathcal{P}}$, if $\mathcal{P}^{\text{joint}}$ were Gaussian distributed, would scale with $\sqrt{\Sigma_{\mathcal{P}^{\text{joint}}}}/N_{\mathcal{R}^{\circ}_{k,t}}$. In practice, we do not know a priori the number of samples in the collision region, and $\mathcal{P}^{\text{joint}}$ is non-gaussian but, compared to an analytical solution, we have empirically observed that we misclassify the CP to be below the threshold less than 2\% of the times.

\subsection{Cost Formulation}
\label{sec:cost}
A key component of MPPI is the design of the cost function $S_k$. We design our cost function as:
\begin{equation}\label{eq:cost}
   S_k =  \mathcal{C}_\text{tracking} + \mathcal{C}_\text{speed}+\mathcal{C}_\text{rotation}+ \mathcal{C}_\text{risk},
\end{equation}
where $\mathcal{C}_\text{tracking}$ drives the robot towards its goal and penalizes the lateral deviation from the reference path, $\mathcal{C}_\text{speed}$ penalizes differences with reference speed, and $\mathcal{C}_\text{rotation}$ penalizes angular velocity. $\mathcal{C}_\text{risk}$ is defined as:
\begin{equation}
\label{eq:mppi_risk_cost}
    \mathcal{C}_\text{risk}= \omega_\text{soft} \hat{\mathcal{P}}(\boldsymbol{q}_{k,t}) + \omega_\text{hard} \mathds{1}_\sigma(\hat{\mathcal{P}}(\boldsymbol{q}_{k,t})),
\end{equation}
with $\omega_\text{soft}$ and $\omega_\text{hard}$ being tunable weights.
The first part, $\omega_\text{soft} \hat{\mathcal{P}}(\boldsymbol{q}_{k,t})$, is a linear penalty on any collision probabilities, incentivising the planner to always choose lower risk.
The second part, $\omega_\text{hard} \mathds{1}_\sigma(\hat{\mathcal{P}}(\boldsymbol{q}_{k,t}))$ is an indicator function returning $1$ when the CP is above a chosen risk threshold $\sigma$.
Every sampled trajectory with a CP above the threshold, assuming that $\omega_\text{hard}$ is high enough and that at least a few low-risk trajectories are sampled, will receive close to zero weight in the average in~\cref{eq:mppi_update}, essentially being rejected.
One advantage is that if, for example, the robot starts planning from a state with CP above the threshold, MPPI still returns a solution in which the CP and the number of time steps with CP above the threshold will be minimised.

\subsection{Algorithm}
\label{sec:algo}
\Cref{alg:dra_mppi} summarizes the proposed approach. Note that operations over $k$ and $j$ are parallelized, while the ones over $t$, i.e. \textit{rolling out} the trajectories, are performed sequentially.

As in \cite{Pezzato}, $\mathcal{E}_k$ are splines fitted to a Halton sequence for improved smoothness, the inverse temperature $\beta$ is updated online, and $U^*$ is time-shifted to be reused in the next iteration. We also always take one sample to be zero velocity throughout the horizon. This helps MPPI to converge to a braking manoeuvre when all other samples fail~\cite{trevisan_biased-mppi_2024}. 

\begin{rmk}
    \label{sec:remark}
A key difference between MPPI and MPC is that MPPI's planned trajectory is \textit{not} meant to be applied in open loop. 
When symmetries are present, e.g. when a robot faces an obstacle ahead, MPPI's plan might go straight through the obstacle. This is intuitively clear given the weighted average in~\cref{eq:mppi_update}.
For certain types of systems and cost functions, MPPI can be related to Path Integral Control~\cite{williams_information-theoretic_2018}.
Within the Path Integral control framework, because there is an assumption that the actions of the ego-agent are noisy, attempting to drive straight to the obstacle \textit{for a while} is the optimal action. The \textit{symmetry} is proven to \textit{break at some point}~\cite{kappen_path_2005}.
For our planner, this means that while we can reject the samples $k$ with collision probability higher than the threshold $\sigma$, we cannot guarantee that the resulting plan doesn't violate the threshold.
Again, this is expected as, in MPPI, only the first input of $U^*$ is applied, and the rest is only used for warm-starting the next iteration.
\end{rmk}

\begin{algorithm}[h]
\caption{Dynamic Risk-Aware MPPI} \label{alg:dra_mppi}

\begin{algorithmic}[1]
\Initialize{$U_{init}=[0,\dots,0]$ \Comment{$U_{init} \in \mathbb{R}^T$} \\
$\mathcal{E}_k \gets \textit{sampleHaltonSplines}()$ \Comment{$k=1...K$}}
\While{$\textit{taskNotDone}$}
\State{$\boldsymbol{x}_0 \gets \textit{observeEnvironment}()$}
\For{$k=1 \dots K$} \Comment{in parallel}
\State{${V}_k = U_{init} + \mathcal{E}_k$}
\For{$t=0 \dots T-1$}
\State{$\boldsymbol{x}_{k,t+1} = \mathcal{F}(\boldsymbol{x}_{k,t}, \boldsymbol{v}_{k,t})$}
\State{$\mathcal{R}^{{\scriptscriptstyle\square}}_t \gets \textit{getArea}(\left[\boldsymbol{x}_{0,t}, ..., \boldsymbol{x}_{K,t}\right])$}   \hfill $\triangleright$ \cref{eq:sampling_region}
\For{$j=1 \dots N_{mc}$} \Comment{in parallel}
\State{$x_j,y_j \gets \textit{sampleUniform}(\mathcal{R}^{{\scriptscriptstyle\square}}_t)$}
\State{$\mathcal{P}^\text{joint}_{j} \gets \textit{evaluateCP}(x_j,y_j)$}   \hfill $\triangleright$ \cref{eq:jointCP_persample}
\EndFor
\State{$\begin{aligned}
\hat{\mathcal{P}}_{k,t+1} \gets & \textit{estimateJointCP}(\boldsymbol{x}_{k,t+1},\\ 
& \left[\mathcal{P}^\text{joint}_{1}, ..., \mathcal{P}^\text{joint}_{N_{mc}}\right])
\end{aligned}$}   \hfill $\triangleright$ \cref{mc}
\EndFor
\State{$S_k \gets \textit{getCost}({X}_k, \left[\hat{\mathcal{P}}_{k,1}, ..., \hat{\mathcal{P}}_{k,T}\right])$}   \hfill $\triangleright$ \cref{eq:cost}
\State{$w_k \gets \textit{importanceSampling}$}   \hfill $\triangleright$ \cref{IS}
\EndFor
\State{$\beta \gets \textit{updateBeta}(\beta, \eta)$}
\State{$U^* = \sum_{k=1}^K w_k {V}_k$}
\State{$U_{init} \gets \textit{timeShift}(U^*)$}
\State{$\textit{applyInput}(\boldsymbol{u}_0^*)$}
\EndWhile
\end{algorithmic}
\end{algorithm}

\begin{figure*}[ht!]
\centering
\medskip
\begin{subfigure}{.33\textwidth}
  \centering
  \includegraphics[width=0.9\linewidth]{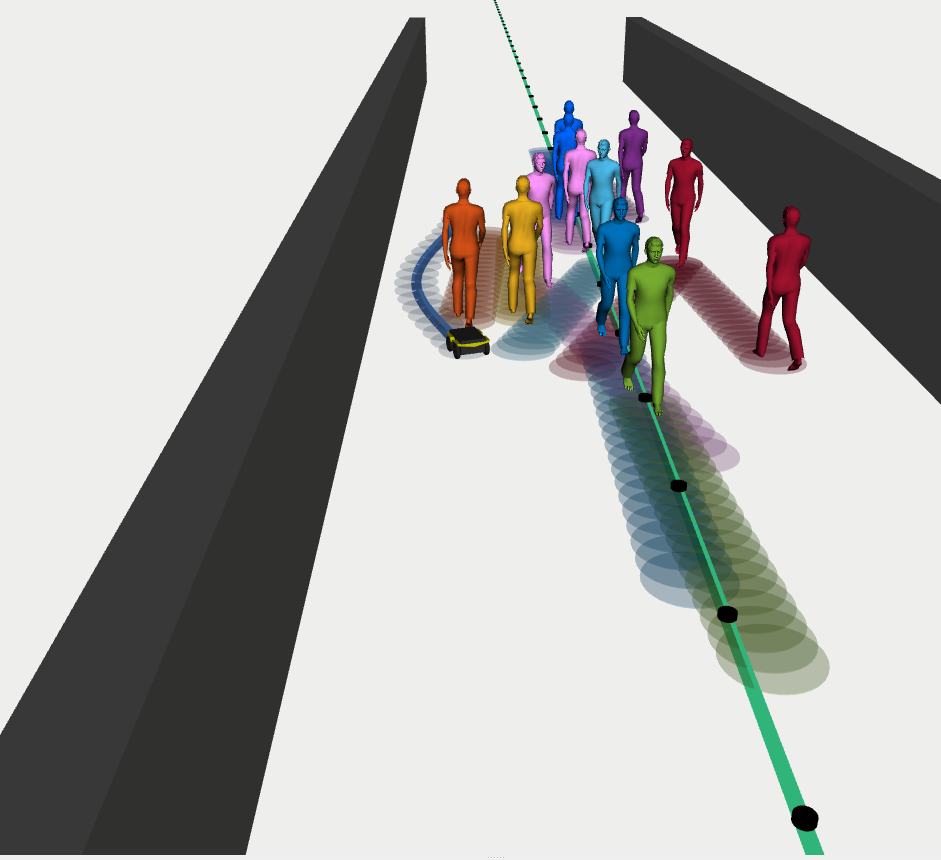}
  \caption{}
  \label{fig:sub1}
\end{subfigure}%
\begin{subfigure}{.33\textwidth}
  \centering
  \includegraphics[width=0.9\linewidth]{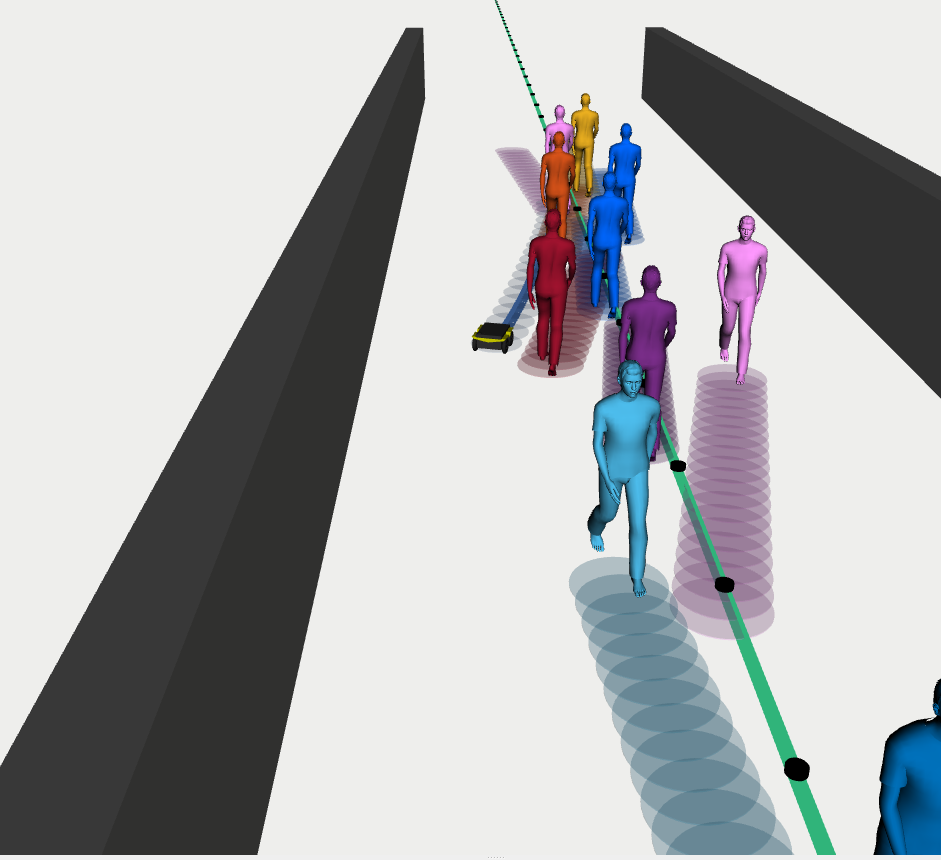}
  \caption{}
  \label{fig:sub2}
\end{subfigure}
\begin{subfigure}{.33\textwidth}
  \centering
  \includegraphics[width=0.9\linewidth]{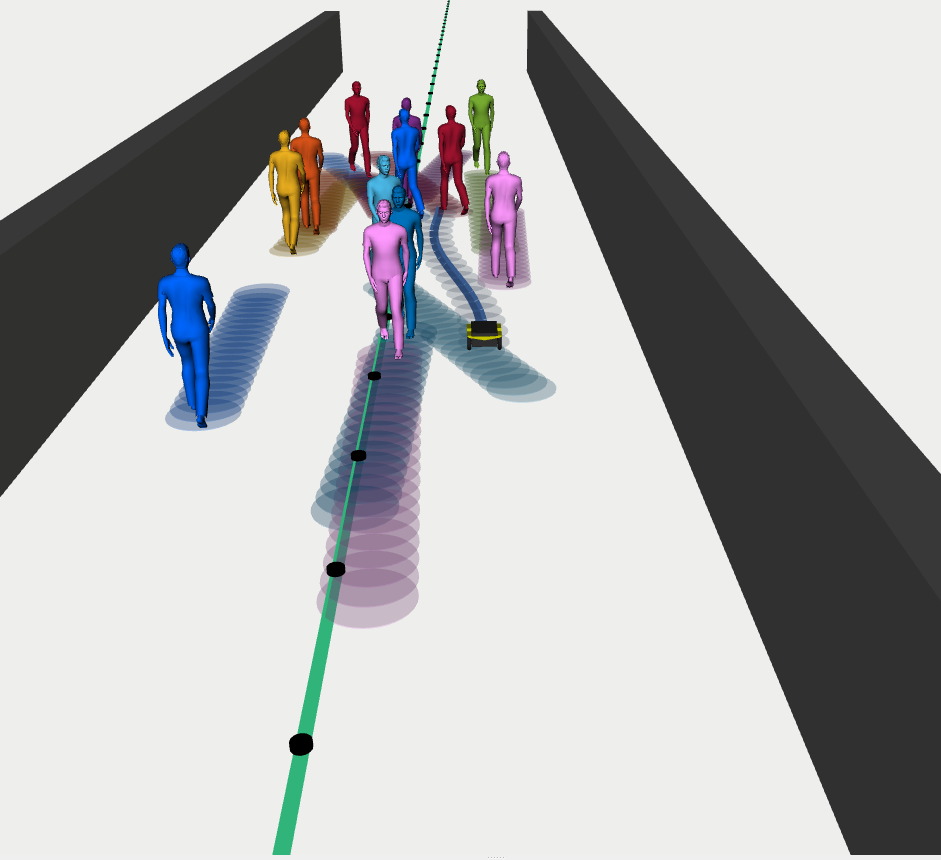}
  \caption{}
  \label{fig:sub3}
\end{subfigure}
\caption{Snapshots from the simulated environment under Gaussian pedestrian motion at different time instants. Blue circles depict the robot's plan, whereas the pedestrians' predicted trajectories are illustrated in distinct coloured circles. The black walls on both sides represent the corridor boundaries.}
\label{fig:simulation}
\end{figure*}

\section{Simulation Experiments}

This section evaluates the proposed approach in the context of mobile robot navigation in a crowded environment shared with humans. Our algorithm's implementation is built upon previous open-source MPPI solvers \cite{Pezzato, trevisan_biased-mppi_2024}. The proposed approach is developed in Python using PyTorch and integrated with the Robot Operating System (ROS). The laptop running the simulations has an Intel\textsuperscript{\textregistered} Core\textsuperscript{TM} i7 CPU@2.6GHz and NVIDIA GeForce RTX 2080.

\subsection{Simulation Setup}
The experimental evaluation is conducted in a simulated environment featuring a Clearpath Jackal robot navigating a 6-meter-wide corridor alongside multiple pedestrians. The robot has to follow the corridor's centerline while maintaining a reference velocity of $2$m/s. A visualization with Gaussian predictions is given in~\Cref{fig:simulation}. The simulator runs at $20$Hz, while the controllers run at $5$Hz.
Pedestrian dynamics are modelled using the Social Forces~\cite{Helbing}, implemented via open-source software~\cite{Gloor}. In this model, pedestrians avoid one another and the robot. Future pedestrian positions are predicted by the robot's motion planner using a constant velocity assumption.
The mismatch between the model used for simulation and prediction is to emulate a real-world scenario where the predictions do not always match the ground-truth trajectory.
Pedestrians are modelled with a radius of $0.3$m and spawn on both sides of the corridor, and their goal is to reach the opposing side. The robot dynamics are described by a second-order unicycle model~\cite{model}. Simulations are conducted for three scenarios featuring $4$, $8$, and $12$ pedestrians to evaluate the planner’s performance under varying crowd density levels.

\subsubsection{Pedestrians with Gaussian noise}
In this setup, we model the uncertainty in pedestrian motion using a unimodal Gaussian distribution with covariance matrix $\boldsymbol{\Sigma}_w =0.3^2 \boldsymbol{I}$. By `unimodal', it is meant that we predict a single trajectory for each pedestrian. We define the pedestrians' dynamics as:
 \begin{equation}
     \boldsymbol{\delta}_{t+1} = \boldsymbol{\delta}t + (\boldsymbol{v} + \boldsymbol{\delta}_{w,t})dt, \quad \boldsymbol{\delta}_{w,t} \sim \mathcal{N}(\boldsymbol{0}, \boldsymbol{\Sigma}_w),
 \end{equation}
 where $\boldsymbol{v} \in \mathbb{R}^2$ is the velocity that follows the social forces model. The additive noise $\boldsymbol{\delta}_{w,t}$ introduces stochastic perturbations, capturing the variability in pedestrian motion.

 \subsubsection{Pedestrians with Mixture of Gaussians} In this setup, pedestrian motion is modelled as a Markov Chain, where transitions govern changes in the movement direction. Specifically, a pedestrian follows a horizontal trajectory with probability $p=0.975$ at each time step while transitioning to a diagonal trajectory with probability $p = 0.025$. This behaviour is further perturbed by Gaussian noise as before. The pedestrians' dynamics are described by:  
\begin{equation}
    \boldsymbol{\delta}_{t+1} = \boldsymbol{\delta}_t + (B\boldsymbol{v} + \boldsymbol{\delta}_{w,t}) dt, \quad \boldsymbol{\delta}_{w,t} \sim \mathcal{N}(\boldsymbol{0}, \boldsymbol{\Sigma}_w),
\end{equation}
\noindent where $B$ is a transformation vector that depends on the current state of the Markov Chain and $\boldsymbol{v}$ is a constant preferred velocity. If the pedestrian is in the horizontal state, $B=\begin{bmatrix} 1 & 0
\end{bmatrix}^T$. Conversely, if the pedestrian switches to the diagonal state, $B = \begin{bmatrix} \nicefrac{1}{\sqrt{2}} & \nicefrac{1}{\sqrt{2}} \end{bmatrix}^T$. The uncertainties associated with this motion can be modelled as a Mixture of Gaussians where each state transition in the Markov Chain leads to a separate mode with an associated probability. For a planning horizon of $T=20$, the pedestrian may switch from horizontal to diagonal movement every 5-time steps, resulting in 4 distinct modes per pedestrian.

\begin{table*}[ht]
    \centering
    \medskip
    \scalebox{1.0}{
    \begin{tabular}{||c |l| c| c| c| c| c||}
        \hline
        \textbf{\# Pedestrians} & \textbf{Method} & \textbf{Task Duration [s]} & \textbf{Velocity [m/s]} & \textbf{CP} & \textbf{SR \%)} &
        \textbf{Runtime [ms]}\\
        \hline
        \hline
        \multirow{4}{*}{4} & SH-MPC \cite{de_groot_scenario-based_2023} & 19.63 (0.98) & 1.79 (0.07) & 0.047 (0.024)& 99 & 53.8 (5.2)\\
        & Frenét-Planner \cite{fiss}  & 20.17 (1.38) & 1.76 (0.11) & 0.112 (0.086) & 91& 105.9 (5.4)\\
        & Vanilla MPPI \cite{williams_information-theoretic_2018} & \textbf{18.81 (0.14)} & \textbf{1.87 (0.01)} & - &  86 & \textbf{48.1 (4.1)}\\
        & DRA-MPPI (ours)  & 19.13 (0.12) & 1.84 (0.04) & 0.020 (0.013) &  \textbf{100}& 96.9 (3.3)\\
        \hline
        \hline
        \multirow{4}{*}{8} & SH-MPC \cite{de_groot_scenario-based_2023} & 20.31 (1.48) & 1.76 (0.11) & 0.051 (0.022) & 96& \textbf{58.6 (5.1)}\\
        & Frenét-Planner \cite{fiss} & 20.88 (1.37) & 1.71 (0.11) & 0.196 (0.084) & 82& 145.9 (9.4)\\
        & Vanilla MPPI \cite{williams_information-theoretic_2018} & \textbf{18.83 (0.23)} & \textbf{1.85 (0.01)} & - & 61 & 58.45 (3.2)\\
        & DRA-MPPI (ours)  & 19.71 (0.26) & 1.82 (0.15) & 0.034 (0.046) & \textbf{98} & 109.9 (7.7)\\
        \hline
        \hline
        \multirow{4}{*}{12} & SH-MPC \cite{de_groot_scenario-based_2023} & 21.34 (1.69) & 1.68 (0.12) & 0.051 (0.065) & 96& \textbf{61.4 (5.4)}\\
        & Frenét-Planner \cite{fiss} & 22.26 (1.55) & 1.61 (0.11) & 0.216 (0.164)  & 74& 171.3 (8.6)\\
        & Vanilla MPPI \cite{williams_information-theoretic_2018}& \textbf{18.98 (0.54)} & \textbf{1.84 (0.03)} & - & 43& 68.64 (3.8)\\
        & DRA-MPPI (ours) & 19.86 (0.14) & 1.78 (0.22) & 0.040 (0.064) & \textbf{98}& 114.2 (5.3)\\
        \hline
    \end{tabular}}
    \caption{Performance comparison of different planning methods under different pedestrian densities. Quantitative results were obtained from over 100 experiments for uni-modal pedestrian predictions. 
    }
    \label{tab:comparison_unimodal}
\end{table*}

\begin{table*}[ht]
    \centering
    \scalebox{1.0}{
    \begin{tabular}{||c |l| c| c| c| c| c||}
        \hline
        \textbf{\# Pedestrians} & \textbf{Method} & \textbf{Task Duration [s]} & \textbf{Velocity [m/s]} & \textbf{CP} & \textbf{SR (\%)} &
        \textbf{Runtime [ms]}\\
        \hline
        \hline
        \multirow{4}{*}{8} & SH-MPC \cite{de_groot_scenario-based_2023} & 20.92 (3.53) & 1.74 (0.18) & 0.053 (0.028) & 95& \textbf{74.8 (6.1)}\\
        & Frenét-Planner \cite{fiss} & 22.26 (3.51) & 1.73 (0.18) & 0.268 (0.146) & 86& 188.9 (6.8)\\
        & Vanilla MPPI \cite{williams_information-theoretic_2018}& 19.71 (0.13) & 1.80 (0.01) & - & 78& 86.41 (4.8)\\
        & DRA-MPPI (ours) & \textbf{19.67 (0.17)} & \textbf{1.81 (0.01)} & 0.024 (0.025) & \textbf{99}& 118.9 (5.3)\\
        \hline
    \end{tabular}}
    \caption{Performance comparison of different planning methods under different pedestrian densities. Quantitative results are obtained from 100 experiments for multi-modal pedestrian predictions.}
    \label{tab:comparison}
\end{table*}

\subsection{Comparison to Baselines}
We compare DRA-MPPI against three baselines. Baselines are selected on the availability of an open-source implementation and their application to navigation in 2-D dynamic
environments. We consider the following baselines:

\begin{enumerate}[(i)]
\item Stochastic Model Predictive Control (SH-MPC) \cite{de_groot_scenario-based_2023}: This baseline uses a scenario-based MPC and formulates the collision avoidance chance constraints over the entire trajectory as a scenario program. 
\item Vanilla MPPI \cite{williams_information-theoretic_2018}: This baseline assumes deterministic pedestrians' behaviour by considering only the mean predicted trajectories and plans around them.
\item Frenét Planner \cite{fiss}: This sampling-based planner generates trajectories in the Frenét coordinate system. The risk metric used is similar to previous work \cite{racp}.
\end{enumerate}
All planners have a horizon of $T = 20$ steps, with a discretization step of $0.2$s, resulting in a time horizon of $4.0$s. We set the maximum allowable risk level to $\sigma = 0.05$. DRA-MPPI uses $K=400$ and $N_{mc}=20000$ samples. The following metrics are considered for evaluation:
\begin{enumerate}
    \item \textit{Task Duration}: Measures the time required for the robot to reach the end of the corridor.
    \item \textit{Velocity}: Determines the average velocity maintained throughout the experiment.
    \item \textit{Collision Probability}: Quantifies each experiment's maximum joint collision probability (CP). The CP we report is computed a posteriori. Given the remark in~\cref{sec:algo}, the CP is only reported for the current time step instead of over the planned trajectory.
    \item \textit{Success Rate (SR)}: Evaluates the proportion of experiments in which the robot successfully avoids collisions with pedestrians and reaches its goal.
\end{enumerate}

\begin{figure}
\centering
\includegraphics[width=0.9\linewidth]{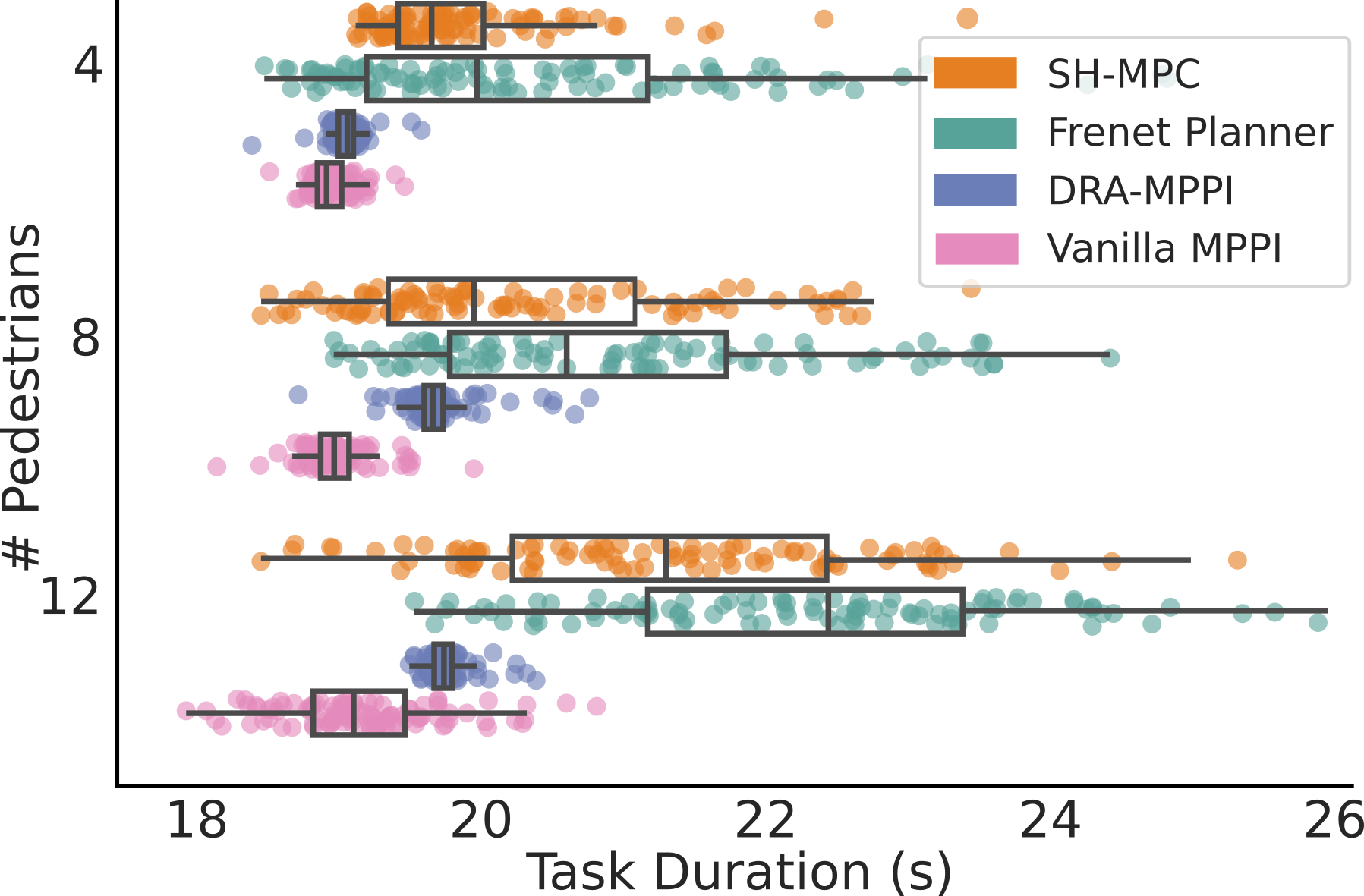}
\caption{The visualization illustrates the task duration, defined as the time required to reach the goal, under a uni-modal pedestrian prediction setting. Our method demonstrates a lower variance in task duration and a shorter completion time compared to SH-MPC and the Frenét Planner. While Vanilla MPPI achieves the shortest task duration, this comes at the cost of a higher collision rate, as reported in Table \ref{tab:comparison_unimodal}.}
\label{boxplot_timeduration}
\end{figure}

\begin{figure}
\centering
\includegraphics[width=0.9\linewidth]{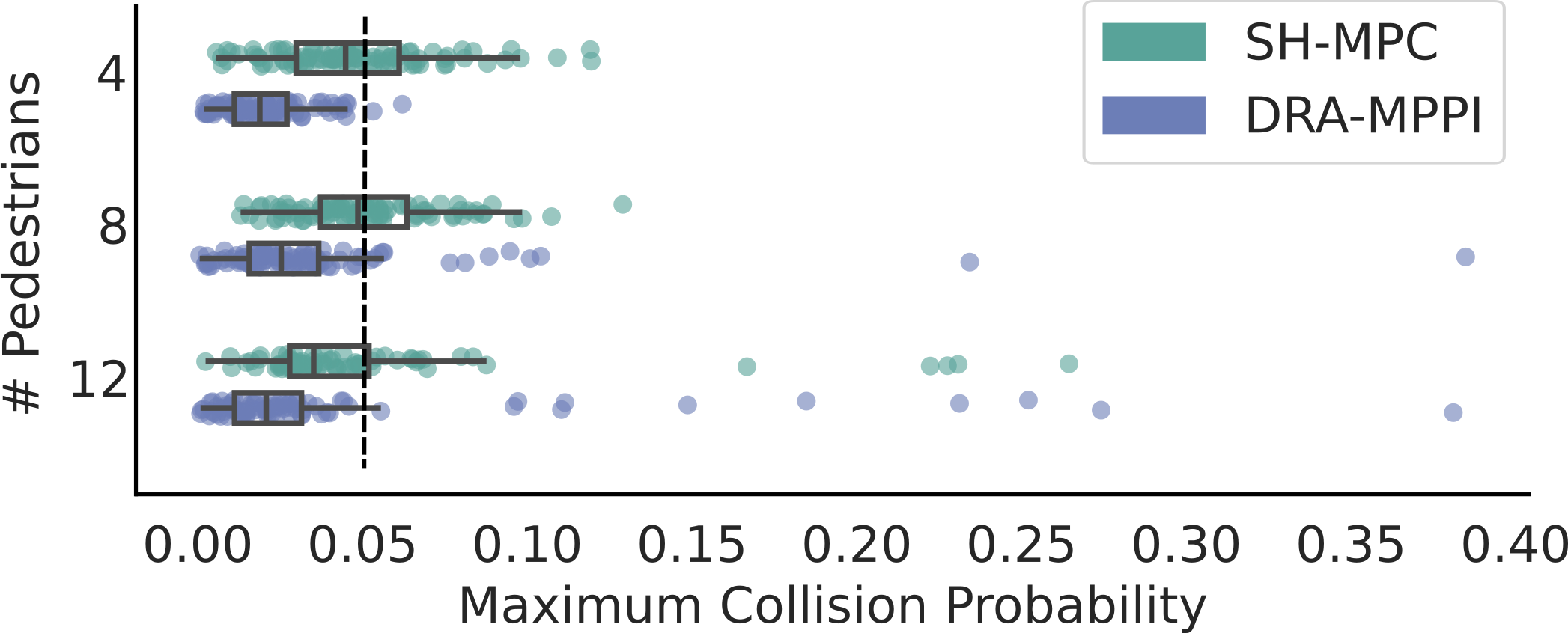}
\caption{Visualization of maximum collision probability across 100 experiments under uni-modal pedestrian prediction setup. Frenét Planner is discarded from this Fig. for better readability as its CP is much higher than SH-MPC and DRA-MPPI as seen in~\Cref{tab:comparison_unimodal}. 
}
\label{boxplot_cp_unimodal}
\end{figure}

\begin{figure}
\centering
\includegraphics[width=0.9\linewidth]{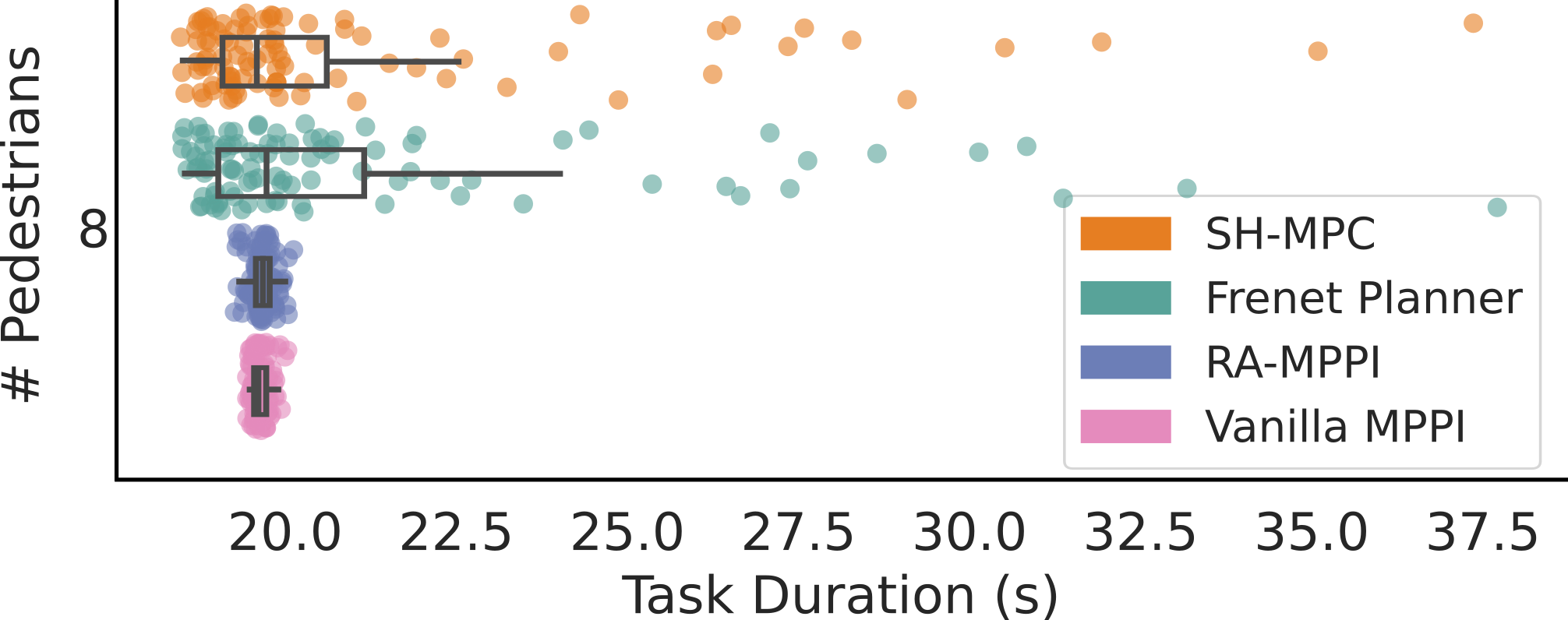}
\caption{Visualization of the task duration under multi-modal pedestrian prediction setting. Same conclusions as the uni-modal case.}
\label{boxplot_timeduration_mm}
\end{figure}

\begin{figure}
\centering
\includegraphics[width=0.9\linewidth]{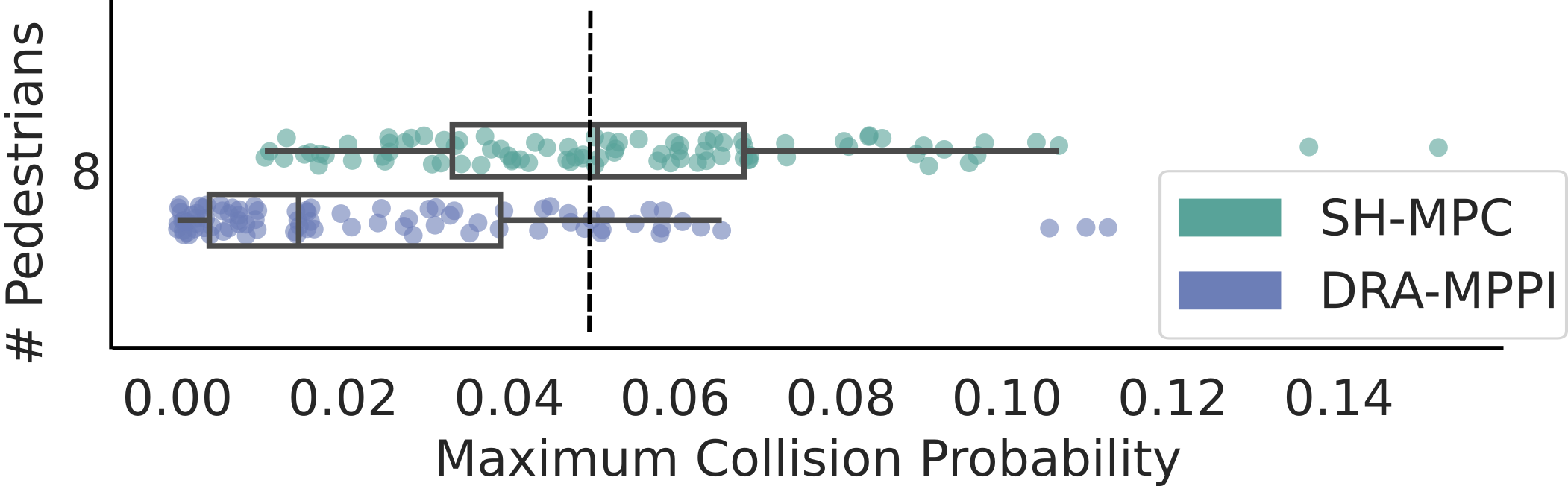}
\caption{Visualization of maximum collision probability across 100 experiments under multi-modal pedestrian prediction setup.}
\label{boxplot_cp_mm}
\end{figure}

\begin{figure*}[h!]
    \medskip
  \centering
  \includegraphics[width=\textwidth]{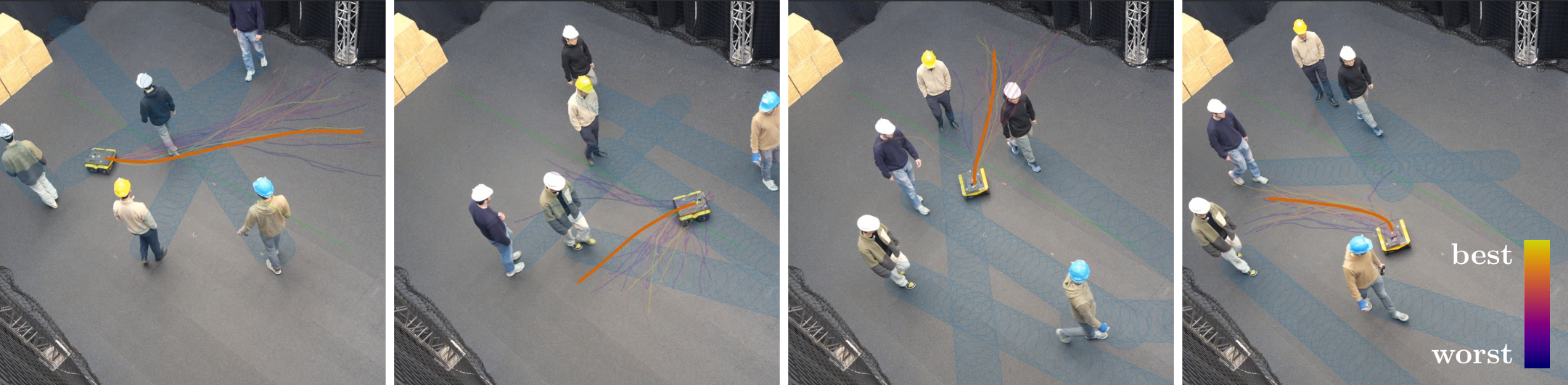}
  \caption{Four snapshots of a Jackal robot running DRA-MPPI among five pedestrians. In orange is the trajectory planned by our proposed approach, with the top 30 samples colour-graded by their relative cost. In blue are the predicted trajectories of the pedestrians, and the circle size represents three standard deviations from the mean predicted centre of the obstacle. In green is the reference path. The behaviour is best appreciated in the accompanying video.}
  \label{fig:pendulum_inputs}
\end{figure*}

\Cref{tab:comparison_unimodal} and \Cref{boxplot_timeduration,boxplot_cp_unimodal} present a quantitative comparison of different planning methods across varying pedestrian densities under uni-modal pedestrian predictions. The results, averaged over $100$ experiments, highlight the trade-offs between task efficiency and safety among the evaluated approaches.
Our proposed approach (DRA-MPPI) consistently demonstrates a favourable balance between task duration and safety.
While Vanilla MPPI achieves the shortest task duration across all scenarios, it does so at the expense of significantly lower safety due to its lack of risk awareness, particularly as pedestrian density increases.
For instance, in the presence of $12$ pedestrians, Vanilla MPPI completes the task in $18.98$s but exhibits the lowest safety rate ($43$\%), indicating a high collision frequency.
In contrast, DRA-MPPI maintains a competitive task duration ($19.86$s) while achieving a significantly higher safety rate of $98$\%, demonstrating its robustness in crowded environments.
The Frenét Planner exhibits longer task duration and suffers a notable drop in safety, particularly at higher pedestrian densities ($74\%$ safety at 12 pedestrians). This deterioration arises from the planner's design, which does not explicitly optimize for risk minimization. Instead, it evaluates a set of sampled trajectories based on assigned costs and selects the one with the lowest one. However, in dense environments, the trajectory with the minimum cost may still entail substantial risk, as observed in the experimental results.
\begin{rmk}
It is important to stress that, in our simulator, the pedestrians move with Social Forces but are predicted by a constant velocity model, i.e. the predictions are not the ground truth behaviour. 
This may result in the planner thinking its plan is below the risk threshold while, at the next iteration, it may find itself at a starting pose with a higher-than-allowed collision probability.
\end{rmk}

SH-MPC strives to track the reference path very closely, often finding itself in riskier positions.
SH-MPC relaxes the constraints with slack variables when no feasible solution is found in these riskier situations. When infeasible, SH-MPC switches to a fallback policy (braking). This leads to SH-MPC taking longer to complete the task and often exhibiting collision probabilities higher than the threshold at least a few times per experiment, severely impacting the metrics.

Thanks to its cost~\cref{eq:cost}, DRA-MPPI always minimizes the CP, even when below the threshold. This makes it strive to avoid riskier situations altogether. Moreover, when DRA-MPPI finds itself in a riskier situation, it will try to return to a state with CP below the threshold in the least amount of steps possible.
For these reasons, DRA-MPPI demonstrates good task completion times while providing the highest safety. Moreover, DRA-MPPI and Vanilla MPPI exhibit a lower variance in task duration across experiments, indicating better consistency than SH-MPC and Frenét Planner.

\Cref{tab:comparison} and \Cref{boxplot_timeduration_mm,boxplot_cp_mm} show that the key observations from the uni-modal case also hold in the multi-modal setting. DRA-MPPI outperforms other methods in balancing task efficiency and safety. It achieves the shortest task duration ($19.67$s) while maintaining the highest safety rate ($99\%$), demonstrating its ability to navigate complex, uncertain environments effectively. In contrast, Vanilla MPPI completes the task in a comparable time ($19.71$s) but at the cost of reduced safety ($78\%$), reflecting its lack of risk-awareness. DRA-MPPI also remains close to the maximum allowable risk while ensuring safety, demonstrating its consistency and robustness in complex pedestrian environments.

\subsection{Computation Time}
In~\Cref{tab:comparison_unimodal,tab:comparison}, we also report the computation time (runtime) for SH-MPC and our approach. DRA-MPPI takes just below $100$ms to compute the control action in the uni-modal four-agents experiment, while SH-MPC is almost twice as fast. 
Both methods scale quite gracefully over the number of agents with multi-modal predictions.
However, it is important to note that SH-MPC is written in C++, while DRA-MPPI is implemented in PyTorch. Recent work shows a reduction of compute time of up to two orders of magnitude when running MPPI in CUDA vs Pytorch~\cite{vlahov2024mppi_generic}, highlighting the potential for these massively parallelizable approaches.

\section{Real Robot Experiments}
This section demonstrates the DRA-MPPI for a mobile robot driving among pedestrians in an arena, highlighting the real-time capability of the proposed approach. 
\subsection{Experimental Setup}
The experiment occurs in a large arena where pedestrians walk alongside the robot. The robot and pedestrians' positions are tracked by a motion capture system operating at 20 Hz. Pedestrian positions are filtered through a Kalman filter that also estimates velocities, which the planner uses to produce constant velocity predictions. The pedestrians are assumed to follow an uni-modal Gaussian distribution with a covariance matrix of $\boldsymbol{\Sigma}_w =0.1^2 \boldsymbol{I}$. The robot follows a reference path between two opposite sides of the environment and performs a turn upon reaching each side. The cost function is the same as in~\cref{eq:cost}. 
\subsection{Results}
In this set of experiments, the pedestrians were instructed to walk naturally from one side to the other while interacting with the robot. The experiment was conducted over a duration of $65$ seconds, with snapshots captured at various time instances, as illustrated in Fig. \ref{fig:pendulum_inputs}. Throughout the trials, the maximum collision probability observed for the robot was $0.054$, with an average of $0.023$ and a standard deviation of $0.011$. Notably, no collisions were recorded.

\section{Conclusions}
\label{sec:conclusions}
We have presented DRA-MPPI, an MPPI-based planner capable of computing, in parallel, at each time step, the joint collision probability among many dynamic obstacles using potentially non-Gaussian predictions.
Thanks to its gradient-free nature, we have shown that DRA-MPPI can reject samples with a collision probability above a desired threshold and directly minimise the collision probability.

In simulated experiments, while the pedestrians were predicted with constant velocity, they moved with Social Forces. DRA-MPPI demonstrated the highest robustness to this mismatch by being the safest method among a Scenario-Based MPC approach, a risk-aware Frenét Planner, and a Vanilla MPPI while maintaining high speed and good task completion times.
Similar conclusions were drawn in experiments where the pedestrians could randomly switch directions, predicted by a Gaussian Mixture Model, demonstrating efficient planning with non-Gaussian distributions.

Real-robot experiments showcased DRA-MPPI's ease of transfer to the real world and real-time performance. Although we relied on a motion capture system, the pedestrian's state and the map can be retrieved from onboard sensors~\cite{brito_mpcc}.

While already real-time, in future work our PyTorch implementation could be rewritten in CUDA, dramatically improving computation times~\cite{vlahov2024mppi_generic}. This could also allow for real-time computations of the Monte Carlo approximation of the joint collision probability over the planning horizon, allowing for accurate constraining of the joint collision probability over the entire trajectory.


\bibliographystyle{IEEEtran}
\bibliography{IEEEabrv, zotero}

\begin{thebibliography}{10}
\providecommand{\url}[1]{#1}
\csname url@samestyle\endcsname
\providecommand{\newblock}{\relax}
\providecommand{\bibinfo}[2]{#2}
\providecommand{\BIBentrySTDinterwordspacing}{\spaceskip=0pt\relax}
\providecommand{\BIBentryALTinterwordstretchfactor}{4}
\providecommand{\BIBentryALTinterwordspacing}{\spaceskip=\fontdimen2\font plus
\BIBentryALTinterwordstretchfactor\fontdimen3\font minus \fontdimen4\font\relax}
\providecommand{\BIBforeignlanguage}[2]{{%
\expandafter\ifx\csname l@#1\endcsname\relax
\typeout{** WARNING: IEEEtran.bst: No hyphenation pattern has been}%
\typeout{** loaded for the language `#1'. Using the pattern for}%
\typeout{** the default language instead.}%
\else
\language=\csname l@#1\endcsname
\fi
#2}}
\providecommand{\BIBdecl}{\relax}
\BIBdecl

\bibitem{amazon}
\BIBentryALTinterwordspacing
M.~Simon. (2019) Inside amazon warehouse where humans and machines become one. [Online]. Available: \url{https://www.wired.com/story/amazon-warehouse-robots/}
\BIBentrySTDinterwordspacing

\bibitem{SD}
\BIBentryALTinterwordspacing
W.~Knight. (2022) Self-driving vehicles are here—if you know where to look. [Online]. Available: \url{https://www.wired.com/story/self-driving-vehicles-here-if-know-where-look/}
\BIBentrySTDinterwordspacing

\bibitem{MI}
\BIBentryALTinterwordspacing
M.~N. Network. (2018) 7 major developments in autonomous shipping in 2018. [Online]. Available: \url{https://www.marineinsight.com/know-more/7-major-developments-in-autonomous-shipping-in-2018/}
\BIBentrySTDinterwordspacing

\bibitem{pfeiffer}
M.~Pfeiffer, G.~Paolo, H.~Sommer, J.~Nieto, R.~Siegwart, and C.~Cadena, ``A data-driven model for interaction-aware pedestrian motion prediction in object cluttered environments,'' \emph{IEEE International Conference on Robotics and Automation}, 2018.

\bibitem{risk-aware}
X.~Huang, A.~Jasour, M.~Deyo, A.~Hofmann, and B.~C. Williams, ``Hybrid risk-aware conditional planning with applications in autonomous vehicles,'' \emph{IEEE Conference on Decision and Control (CDC).}, 2018.

\bibitem{GP}
F.~S. Barbosa, B.~Lacerda, P.~Duckworth, J.~Tumova, and N.~Hawes, ``Risk-aware motion planning in partially known environments,'' \emph{IEEE Conference on Decision and Control}, 2021.

\bibitem{STL}
D.~Sadigh and A.~Kappor, ``Safe control under uncertainty with probabilistic signal temporal logic,'' \emph{Robotics: Science and System}, 2016.

\bibitem{robust-optimization}
A.~Ben-Tal and A.~Nemirovski, ``Robust convex optimisation,'' \emph{Mathematics of Operations Research.}, vol.~23, no.~4, pp. 769--805, 1998.

\bibitem{Freezing}
P.~Trautman and A.~Krause, ``Unfreezing the robot: Navigation in dense, interacting crowd,'' \emph{IEEE/RSJ Int. Conf. on Intelligent Robots and Systems.}, pp. 797--803, 2010.

\bibitem{zhu_chance-constrained_2019}
H.~Zhu and J.~Alonso-Mora, ``\BIBforeignlanguage{en}{Chance-{Constrained} {Collision} {Avoidance} for {MAVs} in {Dynamic} {Environments}},'' \emph{\BIBforeignlanguage{en}{IEEE Robotics and Automation Letters}}, vol.~4, no.~2, pp. 776--783, Apr. 2019.

\bibitem{CC}
A.~Wange, A.~Jasour, and B.~Williams, ``Non-gaussian chance-constrained trajectory planning for autonomous vehicles under agent uncertainty,'' \emph{IEEE Robotics and Automation Letters (RA-L)}, 2020.

\bibitem{de_groot_scenario-based_2021}
O.~De~Groot, B.~Brito, L.~Ferranti, D.~Gavrila, and J.~Alonso-Mora, ``\BIBforeignlanguage{en}{Scenario-{Based} {Trajectory} {Optimization} in {Uncertain} {Dynamic} {Environments}},'' \emph{\BIBforeignlanguage{en}{IEEE Robotics and Automation Letters}}, vol.~6, no.~3, pp. 5389--5396, Jul. 2021.

\bibitem{Okamoto}
K.~Okamoto and P.~Tsiotras, ``Stochastic model predictive control for constrained linear systems using optimal covariance steering,'' \emph{arXiv:1905.13296}, 2019.

\bibitem{Hewing}
L.~Hewing and M.~N. Zeilinger, ``Stochastic model predictive control for linear systems using probabilistic reachable sets,'' \emph{IEEE Conference on Decesion and Control}, 2018.

\bibitem{SP}
M.~C. Campi, S.~Garatti, and F.~A. Ramponi, ``A general scenario theory for nonconvex optimization and decision making,'' \emph{IEEE Transactions on Automatic Control}, vol.~63, no.~12, pp. 4067--4078, 2018.

\bibitem{mustafa}
K.~A. Mustafa, O.~de~Groot, X.~Wang, J.~Kober, and J.~Alonso-Mora, ``Probabilistic risk assessment for chance-constrained collision avoidance in uncertain dynamic environments,'' \emph{IEEE International Conference on Robotics and Automation}, 2023.

\bibitem{Luders}
B.~Luders, M.~Kothari, and J.~How, ``Chance constrained rrt for probabilistic robustness to environmental uncertainty,'' \emph{Proceedings of the AIAA Guidance, Navigation, and Control Conference}, 2010.

\bibitem{Aoude}
G.~S. Aoude, B.~D. Luders, J.~M. Joseph, N.~Roy, and J.~P. How, ``Probabilistically safe motion planning to avoid dynamic obstacles with uncertain motion patterns,'' \emph{Autonomous Robots}, 2013.

\bibitem{nyberg_risk-aware_2021}
T.~Nyberg, C.~Pek, L.~Dal~Col, C.~Noren, and J.~Tumova, ``\BIBforeignlanguage{en}{Risk-aware {Motion} {Planning} for {Autonomous} {Vehicles} with {Safety} {Specifications}},'' in \emph{\BIBforeignlanguage{en}{2021 {IEEE} {Intelligent} {Vehicles} {Symposium} ({IV})}}.\hskip 1em plus 0.5em minus 0.4em\relax Nagoya, Japan: IEEE, Jul. 2021, pp. 1016--1023.

\bibitem{Frenet}
M.~Werling, S.~Kammel, J.~Ziegler, and L.~Groll, ``Optimal trajectories for time-critical street scenarios using discretized terminal manifolds,'' \emph{The International Journal of Robotics Research}, 2012.

\bibitem{williams_information-theoretic_2018}
G.~Williams, P.~Drews, B.~Goldfain, J.~M. Rehg, and E.~A. Theodorou, ``\BIBforeignlanguage{en}{Information-{Theoretic} {Model} {Predictive} {Control}: {Theory} and {Applications} to {Autonomous} {Driving}},'' \emph{\BIBforeignlanguage{en}{IEEE Transactions on Robotics}}, vol.~34, no.~6, pp. 1603--1622, Dec. 2018.

\bibitem{yin_risk-aware_2022}
J.~Yin, Z.~Zhang, and P.~Tsiotras, ``Risk-aware model predictive path integral control using conditional value-at-risk,'' in \emph{2023 IEEE International Conference on Robotics and Automation (ICRA)}.\hskip 1em plus 0.5em minus 0.4em\relax IEEE, 2023, pp. 7937--7943.

\bibitem{caflisch_monte_1998}
R.~E. Caflisch, ``\BIBforeignlanguage{en}{Monte {Carlo} and quasi-{Monte} {Carlo} methods},'' \emph{\BIBforeignlanguage{en}{Acta Numerica}}, vol.~7, pp. 1--49, Jan. 1998.

\bibitem{mppi_williams}
G.~Williams, A.~Aldrich, and E.~A. Theodorou, ``Model predictive path integral control: From theory to parallel computation,'' \emph{Journal of Guidance, Control, and Dynamics}, vol.~40, no.~2, pp. 344--357, 2017.

\bibitem{de_groot_scenario-based_2023}
O.~de~Groot, L.~Ferranti, D.~M. Gavrila, and J.~Alonso-Mora, ``Scenario-based motion planning with bounded probability of collision,'' \emph{The International Journal of Robotics Research}, 2023.

\bibitem{Pezzato}
C.~Pezzato, C.~Salmi, E.~Trevisan, M.~Spahn, J.~Alonso-Mora, and C.~H. Corbato, ``Sampling-based model predictive control leveraging parallelizable physics simulations,'' \emph{IEEE Robotics and Automation Letters}, 2025.

\bibitem{trevisan_biased-mppi_2024}
E.~Trevisan and J.~Alonso-Mora, ``Biased-mppi: Informing sampling-based model predictive control by fusing ancillary controllers,'' \emph{IEEE Robotics and Automation Letters}, 2024.

\bibitem{kappen_path_2005}
H.~J. Kappen, ``\BIBforeignlanguage{en}{Path integrals and symmetry breaking for optimal control theory},'' \emph{\BIBforeignlanguage{en}{Journal of Statistical Mechanics: Theory and Experiment}}, vol. 2005, no.~11, p. P11011, Nov. 2005.

\bibitem{Helbing}
D.~Helbing and P.~Molnár, ``Social force model for pedestrian dynamics,'' \emph{Physical Review E}, no.~5, 1995.

\bibitem{Gloor}
\BIBentryALTinterwordspacing
C.~Gloor, ``Pedsim: Pedestrian crowd simulation,'' 2016. [Online]. Available: \url{https://github.com/chgloor/pedsim}
\BIBentrySTDinterwordspacing

\bibitem{model}
R.~Siegwart and I.~R. Nourbakhsh, ``Introduction to autonomous mobile robots,'' \emph{The MIT Press}, 2011.

\bibitem{fiss}
S.~Sun, Z.~Liu, H.~Yin, and M.~H. Ang, ``Fiss: A trajectory planning framework using fast iterative search and sampling strategy for autonomous driving,'' \emph{IEEE Robotics and Automation Letters}, 2022.

\bibitem{racp}
K.~A. Mustafa, D.~J. Ornia, J.~Kober, and J.~Alonso-Mora, ``Racp: Risk-aware contingency planning with multi-modal predictions,'' \emph{IEEE Transactions on Intelligent Vehicles}, 2024.

\bibitem{vlahov2024mppi_generic}
B.~Vlahov, J.~Gibson, M.~Gandhi, and E.~A. Theodorou, ``Mppi-generic: A cuda library for stochastic optimization,'' \emph{arXiv preprint arXiv:2409.07563}, 2024.

\bibitem{brito_mpcc}
B.~Brito, B.~Floor, L.~Ferranti, and J.~Alonso-Mora, ``Model predictive contouring control for collision avoidance in unstructured dynamic environments,'' \emph{IEEE Robotics and Automation Letters}, vol.~4, no.~4, pp. 4459--4466, 2019.

\end{thebibliography}

\end{document}